\documentclass[letterpaper, 10pt, conference]{ieeeconf}
\usepackage{amsmath}
\usepackage{graphicx}
\usepackage{graphics}
\usepackage{color}
\usepackage{multirow}
\usepackage{algorithm}
\usepackage[noend]{algpseudocode}
\usepackage[export]{adjustbox}

\title{\LARGE \bf qRRT: Quality-Biased Incremental RRT \\for Optimal Motion Planning in Non-Holonomic Systems}

\author{
Nahas Pareekutty$^1$, Francis James$^1$, Balaraman Ravindran$^2$, Suril V. Shah$^1$
}\IEEEoverridecommandlockouts

\graphicspath{{}{Plots/}}

\begin{document}

\maketitle
\IEEEpeerreviewmaketitle
\pagestyle{empty}
\thispagestyle{empty}

\footnotetext[1]{Robotics Research Centre, International Institute of Information Technology - Hyderabad, India}
\footnotetext[2]{The Department of Computer Science and Engineering, Indian Institute of Technology Madras, India}

\begin{abstract}
This paper presents a sampling-based method for optimal motion planning in non-holonomic systems in the absence of known cost functions. It uses the principle of 'learning through experience' to deduce the cost-to-go of regions within the workspace. This cost information is used to bias an incremental graph-based search algorithm that produces solution trajectories. Iterative improvement of cost information and search biasing produces solutions that are proven to be asymptotically optimal. The proposed framework builds on incremental Rapidly-exploring Random Trees (RRT) for random sampling-based search and Reinforcement Learning (RL) to learn workspace costs. A series of experiments were performed to evaluate and demonstrate the performance of the proposed method.
\end{abstract}

\section{Introduction} \label{intro}
\noindent

Optimality is a desired feature in all robot motion planning tasks due to requirements such as minimizing fuel consumption, reducing wear, and eliminating risks. Robot complexities like non-holonomic constraints and workspace complexities like obstacles, terrain and narrow paths make optimal path planning increasingly difficult and is therefore of particular research interest. The motivation for this paper was to develop an optimal motion planning method for non-holonomic systems. \par
Random search methods are good candidates for finding solution paths in complex, high-dimensional systems. Tree-based methods like RRT \cite{rrt} were observed to have the advantage of rapidly generating feasible solution paths while maintaining motion constraints \cite{kinorrt}. However, with these methods, finding an optimal solution is highly unlikely \cite{rrt-subopt}. \par
In systems where cost of actions in the configuration space is either known or calculable, optimized random search methods like A* \cite{a*}, BIT* \cite{bit*} are guaranteed to converge to an optimal solution. The requirement for prior knowledge of cost function makes this method unsuitable for systems in which total trajectory costs are calculable, but local cost functions are not available. \par
Methods like RRT* \cite{rrt*} rely on rewiring connections to trace low cost paths and achieve asymptotic optimality. However, rewiring is only possible in systems with simple dynamics. In non-holonomic systems, rewiring connections is not always feasible due to the presence of differential constraints. Since the guarantee of optimality in these methods depends on rewiring, optimality is lost upon removing this step. For similar reasons, route planning in graph based methods such as PRM \cite{prm} and BIT* are also not a viable option for the systems under our consideration. Other popular methods like RRT-LQR \cite{rrt-lqr} and Kinodynamic RRT* \cite{kinorrt} were also analyzed, but approximations using linearization made them incompatible with complex non-holonomic systems like the Reactionless Manipulation system described in Sec.\ref{exp:reactless}. \par
In the absence of a cost function, \textit{informed} and heuristically-guided RRT methods showed promise. Heuristically-Guided RRT (hRRT) \cite{hrrt} proposes biasing Voronoi regions based on the estimated cost of the region. Such a biasing would guide the RRT growth towards the lower cost regions of the workspace. Other heuristically-guided methods, such as Anytime RRT \cite{anyrrt} and RRT-PI \cite{rrtpi}, begin with an approximate cost function that is iteratively improved with every query. Anytime RRT uses heuristics to restrict the area explored by the random search. Here, the cost function provides an estimate of the value/cost-to-go of each extension. Using this value, potential high cost sub-paths are avoided, leading to greater exploration among the low cost regions of the workspace. RRT-PI proposes replacement of the Euclidean distance used in nearest node search with a value-based measure. Here, the nodes of the solution tree provided by RRT are evaluated using Reinforcement Learning (RL) methods. These nodes are used to estimate the value of new nodes in subsequent queries. The use of RL and a value-based distance measure not only eliminates the need for a known cost function, but also promotes iterative improvement of the cost estimate. \par
A drawback of both Anytime RRT and RRT-PI is that search in the configuration space is biased towards the initial solution. In systems where multiple homotopic solutions exist, these methods may fail to find the alternate solution(s). Also, in Anytime RRT, the performance is dependent on the choice of heuristic, which may not be known for some complex systems. For non-holonomic systems, the path planner should employ a random sampling-based search method along with a heuristic that can independently estimate the region cost. \par
We propose Quality-Biased RRT (qRRT), which combines the rapid exploration and state coverage completeness of incremental RRT with probabilistic sampling of actions and the experience-based method of Reinforcement Learning (RL) to learn region costs. To the best of the authors' knowledge, this approach has not been previously reported and forms the fundamental contribution of this paper. The sample trajectories generated by RRT are evaluated using RL and this information is used to bias the search towards low-cost regions of the configuration space. The coverage completeness of incremental RRT and iterative cost update using RL can be shown to satisfy conditions for asymptotic optimality. Combining these principles in qRRT results in the following advantages:
\begin{itemize}
	\item Coverage completeness of configuration space implies that all feasible states are sampled with a probability of one as the number of samples approach infinity.
	\item Probablistic sampling of actions ensures completeness in action coverage.
	\item RL-based acquisition of cost information eliminates need for prior knowledge of cost information.
	\item Greedy policy using cost information is asymptotically optimal.
	\item Unlike graph-based methods like RRT*, constraints of non-holonomic systems can be maintained without approximations.
	\item Separate random and biased search operations ensure that explorative and exploitative actions are independent.
\end{itemize}

The remainder of this paper is organized as follows: Section \ref{preliminaries} briefly describes the background of various methods used in this paper, Section \ref{evolution} describes motivating factors behind the design of qRRT, Section \ref{algorithm} details the qRRT algorithm, Section \ref{experiments} presents the experimental results and Section \ref{conclusion} discusses the observations made and conclusions. \par

\section{Preliminaries} \label{preliminaries}
\noindent
The qRRT method uses the principles of single-query incremental RRT and RL to generate motion trajectories. The cost (or value) function and policy are approximated using Adaptive Neural Networks. These fundamentals are briefly described in this section. \par

\subsection{Rapidly-exploring Random Trees} \label{secrrt}
\noindent
The RRT method and its algorithm are detailed in \cite{rrt}. This sub-section provides a brief outline of the method. \par 
Path planning in RRT basically consists of a tree that grows from the start node, exploring the configuration space. Each node in the tree is a state within the configuration space. Branching or \textit{extend} operation involves the transition from an existing node in the tree ending in a new node. This extend operation is performed by first selecting a random target state within the configuration space. Then, the node on the tree closest to the target state is selected and transitions in the direction of the target are calculated. Such a transition has two advantages. Firstly, the extend operation favors unexplored areas in the configuration space. Secondly, any motion-related constraints can be included in the transition. These constraints could be motion constraints of the system itself or constraints introduced by obstacles in the workspace. The tree grows through the continuous branching process and the growth is terminated when one of the nodes reaches the goal node. \par
The growth of the tree can be influenced by \textit{biasing} regions within the search space. This helps in guiding the search towards low-cost regions or the goal. \par
As discussed earlier in Section \ref{intro}, RRT is guaranteed to produce a sub-optimal result. To improve the results obtained from RRT, continuously searching single-query variants (\cite{rrt*} and \cite{infrrt}) were introduced in which the search is not terminated when goal is reached. Instead, new nodes are generated and the tree grows until termination conditions like maximum execution time, minimum cost improvement threshold etc. The continuous search ensures that the wider areas of the workspace are covered in search of lower cost paths. \par

\subsection{Reinforcement Learning} \label{RL}
\noindent
Reinforcement Learning is a learning method that uses interaction with the system to learn its characteristics \cite{rl}. The system is modeled as a Markov Decision Process (MDP) consisting of states and actions. The states describe the status of the system at any point of time and actions determine the transitions from one state to other. The effectiveness of an action is defined by its \textit{quality} which includes both the immediate result of the action and the expected consequences further ahead in the future. The immediate result is alternatively defined as a \textit{reward} and the future costs are defined as the \textit{value} of the new state created by the action. Since a state can be followed by multiple actions, the value (or cost-to-go) of a state is determined by the quality of the various actions and the probability of performing each of those actions. \par
Consider a state $s$ and an action $a$ permissible from state $s$, resulting in a new state $s'$. The mapping between states and actions is given by policy $\pi$. The value of the state $s$ while following policy $\pi$ is given by the Bellman equation,
\begin{equation} \label{bellman_value}
V^\pi(s) = \sum\limits_{a} \pi(s,a) \sum\limits_{s'} P^a_{ss'} [ R^a_{ss'} + \gamma V^\pi(s') ]
\end{equation}
where $\pi(s,a)$ gives the probability of choosing action $a$ at state $s$ as defined by the policy, $P^a_{ss'}$ is the probability of transitioning to state $s'$ from $s$ on performing action $a$, $R^a_{ss'}$ is the reward for the transition and $\gamma$ is the discount factor. Future rewards are accumulated using the calculated value of the next state, $V^\pi(s')$. \par

The definition of an optimal control policy $\pi^*$ would be one that selects actions that maximize returns for all states. The Bellman equation for the optimal policy is given by 
\begin{equation} \label{bellman_opt_value}
V^*(s) = \max_a \sum\limits_{s'} P^a_{ss'} [ R^a_{ss'} + \gamma V^*(s') ]
\end{equation}

It is evident that deduction of the optimal policy would require knowledge of the transition probabilities of the system. Also, since the equation is recursive, the value depends on the accumulated rewards of all future states. Calculation of the value would be therefore delayed until all subsequent states till the end of the episode are evaluated. To avoid this delay, the value of a state $s$ can be updated based on the immediate reward and an estimate of the value of the next state $s'$. Such methods are called Temporal Difference (TD) updates. The equation for the simplest TD method, TD(0), is given by 
\begin{equation} \label{td_value}
V(s_t) = (1-\eta)V(s_t) + \eta [ R_t + \gamma V(s_{t+1}) ]
\end{equation}
where $V(s_t)$ and $V(s_{t+1})$ are the estimates of the value of states at time $t$ and $t+1$ respectively. $R_t$ is the immediate reward and $\eta$ is the step-size parameter. \par
It is shown in \cite{rl} that if $\eta$ has a small value, the value function converges with increasing number of update samples. The control policy can be modified to reflect the changes in state values after each value function update. In Policy Iteration, value function updates and policy updates are performed alternatively and through this iterative process, the policy converges towards $\pi^*$. \par

\subsection{Function Approximation using Adaptive Neural Networks} \label{aann}
Use of multi-layer artificial neural networks as approximations of functions has been described in \cite{nn}. Adaptive Artificial Neural Networks (AANNs) can also be designed to continuously update the input-output relations based on new information. Such an approach is suited to sampling-based systems in which data is not available beforehand, but is incrementally acquired as samples. The adaptive nature of AANNs is an advantage over methods like Support Vector Regression. Also, since the information is encoded in the network weights after training, AANNs are independent of the number of samples and is therefore faster and more memory-efficient compared to methods like K-Nearest Neighbors. \par

\noindent
\section{Evolution of Design} \label{evolution}
\noindent

Quality-Biased RRT was developed as a solution for optimal path planning for manipulator robots in space. Since these robots are mounted on a service satellite, movements of their arms cause a reaction in the base satellite operating in free-floating mode. Such reactions may result in loss of attitude which must be corrected or avoided. Avoiding rotations due to arm actions can be ensured by performing actions that are mathematically derived to transmit a net zero reaction to the base satellite. This is commonly referred to as Reactionless Manipulation. Being a non-holonomic system, this placed a lot of constraints on the planner. \par
Based on our analysis of the system and previous research, the requirements identified for such a planner are listed below:
\begin{itemize}
	\item Probabilistically complete explorative search of states and actions
	\item Automatic learning of values function and greedy policy
	\item Path planning using exploitative search 
	\end{itemize}

Implementation details of these requirements are detailed in the following sub-sections. \par

\subsection{Probabilistically Complete Explorative Search of States and Actions} \label{prob_comp}
\noindent
To achieve optimality described by Eq. \eqref{bellman_opt_value}, the search process should ensure complete state coverage through sampling and evaluate all possible transitions/sub-paths from each state. State coverage can be guaranteed through the probabilistic completeness of RRT in a single-query incremental implementation. However, action coverage completeness is not guaranteed in RRT. To achieve this, nodes are also created using random actions performed from the states already present in the tree, as shown in Fig. \ref{fig:block}. Probabilistic sampling of these random actions ensure that all actions will be sampled asymptotically. \par

\subsection{Automatic Learning of Values Function and Greedy Policy}
\noindent
Since we assume that the cost/value function is not available for the systems considered in this paper, this function has to be learned through experience. This is achieved using the Reinforcement Learning process described in Section \ref{RL}. Exploratory actions are executed through the random extend operations performed in RRT and the random actions described in Section \ref{prob_comp}. Exploitation is performed through periodic execution of the greedy policy during the tree growth. \par
An episode is completed whenever a new node is created sufficiently close to the goal state. At the end of each episode, the nodes in the solution path from the start node to the goal state are evaluated. The value update of each node state is performed using the TD update described in Eq. \eqref{td_value}. \par
To identify greedy actions, the samples are grouped based on their state and then the action that results in the highest value is selected as the greedy action. This is repeated at the end of each episode. While state-value updates are performed only for each solution path, the greedy policy is updated by evaluating the state groups in the entire tree. This ensures that greedy actions are updated even when the value landscape changes. \par
Since this method is sampling-based, generalization is required to produce estimates for unsampled regions of the configuration space. This generalization is performed using AANNs. The AANN for the value function accepts the state as input and presents the state value as output. At the end of each episode, this AANN is retrained using the new state-value updates using the back-propagation algorithm. Similarly, the greedy policy is also encoded within an AANN which outputs the greedy action for a given input state. The policy AANN is trained with the new state-greedy action updates at the end of each episode. \par
Encoding the value and policy within AANNs also aids in Transfer Learning between similar systems. AANNs trained within one system can be used as prior knowledge in a similar system and adapted accordingly. Such a learning transfer can significantly reduce exploration time. \par

\subsection{Path planning using Exploitative Search}
\noindent
After completion of every episode, the greedy policy is evaluated by executing it from the start node. This process begins at the start state and involves execution of the greedy action, obtained from the policy AANN, at every step of the trajectory. If the trajectory generated by the greedy policy has greater returns than the current solution, the solution is updated with this trajectory. Convergence to the optimal policy can also be achieved by gradual reduction of exploration and increase of exploitation as the number of samples increases. Exploitation is also performed by intermittently selecting the greedy action during the search process. This helps bias the tree growth towards the high value regions of the configuration space, thereby speeding up the search process. \par
It should be noted here that the exploitative actions performed here replace the rewiring steps of RRT* by tracing a new branch through low cost regions. Unlike the rewiring step, no approximations or simplifications are performed here. Instead, the greedy actions are directly performed on the model, ensuring that the non-holonomic motion constraints are followed in all transitions within the tree. \par

\section{Algorithm} \label{algorithm}
\noindent

\begin{figure}[ht] 
	\centering
	\includegraphics[width=0.5\textwidth]{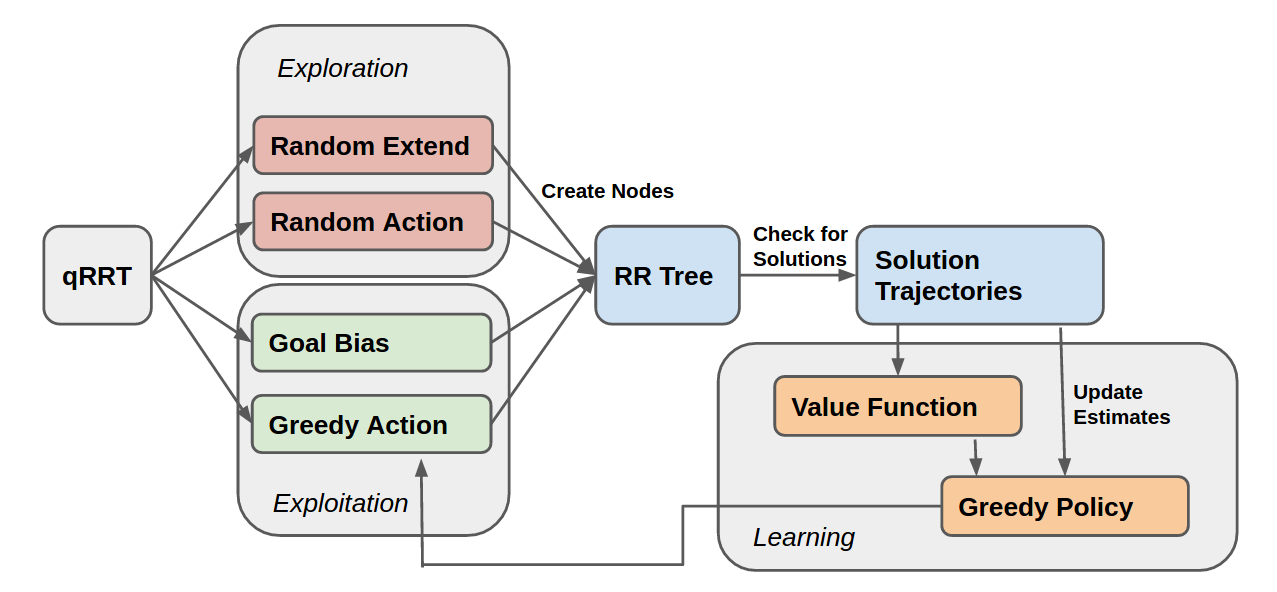}
	\caption{\small{Block Diagram of qRRT}}
	\label{fig:block}
\end{figure}

This section presents the qRRT algorithm which is the fundamental contribution of this paper. The qRRT algorithm executes a continuously growing RRT until the termination condition has been met. In addition to the extend operations of a normal RRT implementation, the qRRT algorithm also includes goal-biased, quality-biased and random action extend operations. The algorithm of qRRT can be broadly divided into: Exploration, Learning and Exploitation, as shown in Fig. \ref{fig:block}. \par

\vspace{-5pt}
\vspace{-5pt}

\subsection{Tree-based Search} \label{main_func}

\begin{algorithm}[ht]
	\caption{\textbf{qRRT}()}\label{qrrt}
	\begin{algorithmic}[1]
		\State \textit{rrTree}.\textbf{add}( INIT\_NODE )

		\State \textit{stateValNN} $\gets$ \textbf{createNeuralNet}()
		\State \textit{optActionNN} $\gets$ \textbf{createNeuralNet}()
									
		\While {\textit{termCondition} $!=$ TRUE}
			\State \textit{sampleType} $\gets$ \textbf{genSampleType}()
			\State \textit{newNode} $\gets$ \textbf{createNode}( \textit{sampleType} )
			\State \textit{rrTree}.\textbf{add}( \textit{newNode} )
			\If {\textbf{reachedGoal}( \textit{newNode} )}
				\State \textit{endNodes}.\textbf{add}( \textit{newNode} )
				\State \textbf{updateStateValues}( \textit{stateValNN}, \textit{newNode} )
				\State \textbf{updatePolicy}( \textit{optActionNN}, \textit{stateValNN} )
			\EndIf

			\State \big[ \textit{maxTraj}, \textit{maxVal} \big] $\gets$ \textbf{getMaxTraj}( \textit{endNodes} )
			\State \big[ \textit{greedyTraj}, \textit{grdVal} \big]  $\gets$ \textbf{getGreedyTraj}( \hfill \\ 
											\hfill \textit{optActionNN} )

			\If {\textit{grdVal} $>$ \textit{maxVal} ) }
				\State \textit{qSolTraj} $\gets$ \textit{greedyTraj}
			\Else
				\State \textit{qSolTraj} $\gets$ \textit{maxTraj}
			\EndIf
		\EndWhile
		\State \Return \textit{qSolTraj}
	\end{algorithmic}
\end{algorithm}

The \textbf{qRRT} function (alg.\ref{qrrt}) manages the growth of the tree (\textit{rrTree}) and the output of solution trajectories. Tree growth is performed through explorative and exploitative transitions that result in the creation of new nodes. The AANNs for value function and policy are \textit{stateValNN} and \textit{optActionNN} respectively. In each iteration, a new node is created using the \textbf{createNode} function and added to \textit{rrTree}. The type of extend operation used to create the node is determined by the sample type returned by \textbf{genSampleType}. This function samples from the available sample types based on their individual sampling probabilities. The extend operations executed by the \textbf{createNode} function are \textbf{extRandState}, \textbf{extGoalState}, \textbf{extRandAction} and \textbf{extGreedyAction}. \par
If the new node is sufficiently close to the goal node, the complete path leading to it from the start node is considered as one of the solution trajectories. The \textbf{updateStateValues} (alg.\ref{updatestatevalues}) function traces the new solution trajectory and updates \textit{stateValNN} with the values of states visited in the trajectory. The greedy policy in \textit{optActionNN} is updated based on the modified \textit{stateValNN} using \textbf{updatePolicy} (alg.\ref{updatepolicy}). \par
The solution trajectory generated in each iteration is chosen from the trajectories within \textit{rrTree} and the trajectory generated by following the greedy policy. The \textbf{getMaxTraj} evaluates all trajectories ending at the goal node and returns the trajectory with the highest quality (or lowest cost). The \textbf{getGreedyTraj} creates a trajectory by sequentially performing the greedy action from the start node until the goal node is reached. The iteration solution trajectory is the one with the highest quality. The \textbf{qRRT} solution trajectory is the last generated iteration solution at the time of termination.\par

\subsection{Exploration}

\begin{algorithm}
	\caption{\textbf{extRandState}()}\label{extrandstate}
	\begin{algorithmic}[1]
		\State \textit{randState} $\gets$ \textbf{genRandState}()
		\State \textit{parentNode} $\gets$ \textbf{getClosestNode}( \textit{randState} )
		\State \big[ \textit{newState}, \textit{newAction} \big] $\gets$ \textbf{performTransition}( \hfill \\ 
											\hfill parentNode, randState )
		\State \textit{transCost} $\gets$ \textbf{getTransCost}( parentNode, newState )
		\State \textit{newNode} $\gets$ \textbf{createNodeObj}( \textit{newState}, \textit{randAction}, \hfill \\ 
											\hfill \textit{transCost} )
		\State \Return \textit{newNode}
	\end{algorithmic}
\end{algorithm}

The random extend operation (\textbf{extRandState}) and random action operation (\textbf{extRandAction}) form the explorative actions of qRRT. The generic extend operation of RRT is implemented in \textbf{extRandState} (alg.\ref{extrandstate}). A random state is selected using \textbf{genRandState} and the closest node on the tree is found using \textbf{getClosestNode}. The \textit{distance} between states has a system-specific definition and may be a direct transition distance or an approximation. Similarly, the \textbf{performTransition} is also a system-specific steering function which returns the steering action and the new state. If a steering function is not available, an approximate heuristic can be used here and can be retrained based on samples generated in \textbf{extRandState} and \textbf{extRandAction}. The transition cost is calculated by \textbf{getTransCost} using local costs/rewards. The new node created consists of state, action and cost information. \par

The \textbf{extGoalState} function is similar to \textbf{extRandState}, with the only difference being the choice of the goal node as steering target instead of a random node. \par

\begin{algorithm}
	\caption{\textbf{extRandAction}()}\label{extrandaction}
	\begin{algorithmic}[1]
		\State \textit{parentNode} $\gets$ \textbf{getRandNode}()
		\State \textit{randAction} $\gets$ \textbf{genRandAction}()
		\State \textit{newState} $\gets$ \textbf{performAction}( parentNode, randAction )
		\State \textit{transCost} $\gets$ \textbf{getTransCost}( parentNode, newState )
		\State \textit{newNode} $\gets$ \textbf{createNodeObj}( \textit{newState}, \textit{randAction}, \hfill \\ 
											\hfill \textit{transCost} )
		\State \Return \textit{newNode}
	\end{algorithmic}
\end{algorithm}

The goal of the \textbf{extRandAction} function (alg.\ref{extrandaction}) is to ensure action coverage completeness. A random node on the tree is picked using \textbf{getRandNode} and a random action is generated using \textbf{genRandAction}. The random action can be a member of the set of permissible actions or a member of the sub-set of actions permissible from the random state. The \textbf{performAction} function executes the transition and subsequently, a new node is created. \par

\subsection{Learning}

The learning process within qRRT involves the training of \textit{optActionNN} and \textit{stateValNN} based on the samples acquired during tree growth. The \textit{stateValNN} AANN is updated whenever a new solution trajectory is found. In \textbf{updateStateValues} (alg.\ref{updatestatevalues}), TD updates are calculated based on Eq. \eqref{td_value}. These updates are stored as state-value pairs in \textit{trainSamples} and used as a batch to retrain \textit{stateValNN}. \par

\begin{algorithm}
	\caption{\textbf{updateStateValues}( \textit{stateValNN}, \textit{endNode} )} \label{updatestatevalues}
	\begin{algorithmic}[1]
		\State \textit{nextStateVal} $\gets$ GOAL\_REWARD
		\While { \textit{endNode}.\textbf{parent}() $<>$ 0 }
			\State \textit{currStateVal} $\gets$ \textit{stateValNN}.\textbf{value}( \textit{endNode}.\textbf{state}() )
			\State \textit{updateStateVal} $\gets$ (1-$\eta$) currentValue $+$ \hfill \\ 
										\hfill $\eta$ \textit{nextStateVal}  
			\State \textit{trainSamples}.\textbf{add}( \textit{endNode}.\textbf{state}(), \textit{updateStateVal} )
			\State \textit{nextStateVal} $\gets$ \textit{endNode}.\textbf{transCost} $+$ \hfill \\
										\hfill $\gamma$ \textit{currStateVal}
			\State \textit{endNode} $\gets$  \textit{endNode}.\textbf{parent}()
		\EndWhile
		\State \textit{stateValNN}.\textbf{train}( \textit{trainSamples} ) 
	\end{algorithmic}
\end{algorithm}

The greedy policy is encoded within \textit{optActionNN} by training it with greedy actions at each state. To do this, sampled states are ordered by their parent nodes and then grouped based on their proximity to each other. Each group consists of transitions starting from a similar state and ending in new state based on the actions performed. The greedy action is selected by choosing the action in each group that produced the highest improvement in value. These steps are performed in \textbf{updatePolicy}. \par

\begin{algorithm}
	\caption{\textbf{updatePolicy}( \textit{optActionNN}, \textit{stateValNN} )} \label{updatepolicy}
	\begin{algorithmic}[1]
		\State \textit{stateActionList} $\gets$ \textbf{sortByParent}
		\State \textit{stateGroups} $\gets$ \textbf{groupByNeigh}( \textit{stateActionList} )
		\For {each \textit{stateGroup} in \textit{stateGroups} }
			\For {each \textit{stateAction} in \textit{stateGroup} }
				\State \textit{stateValue} $\gets$ \textit{stateValNN}.\textbf{value}( \hfill \\ 
										\hfill \textit{stateAction}.\textbf{state}() )
				\State \textit{sTransCost} $\gets$ \textit{stateAction}.\textbf{transCost}() 
				\State \textit{actionQuality} $\gets$ \textit{stateValue} $+$ \textit{sTransCost} 
				\If { \textit{actionQuality} $>$ \textit{maxQuality} }  
					\State \textit{maxQuality} $\gets$ \textit{actionQuality} 
					\State \textit{maxAction} $\gets$ \textit{stateAction}.\textbf{action}() 
				\EndIf
			\EndFor
			\State \textit{trainSamples}.\textbf{add}( \textit{stateGroup}.\textbf{state}(), \textit{maxAction} )
		\EndFor

		\State \textit{optActionNN}.\textbf{train}( \textit{trainSamples} ) 
	\end{algorithmic}
\end{algorithm}

The greedy actions are stored as state-action pairs in \textit{trainSamples} and used for batch-wise retraining of \textit{optActionNN}. \par

\subsection{Exploitation}

Exploitative actions are performed using greedy actions (\textbf{extGreedyAction}) and evaluating the greedy trajectory (\textbf{getGreedyTraj}). The \textbf{extGreedyAction} function (alg.\ref{extgreedyaction}) selects a random state on the tree and performs the action given by the greedy policy AANN, \textit{optActionNN}, for that state. This implements the quality-biased extend operation. The function can be modified to generate a sequence of quality-biased extend operations as long as values increase and goal node is not reached. This can aid in rapid exploitation of acquired information. \par

\begin{algorithm}
	\caption{\textbf{extGreedyAction}()}\label{extgreedyaction}
	\begin{algorithmic}[1]
		\State \textit{parentNode} $\gets$ \textbf{getRandNode}()
		\State \textit{greedyAction} $\gets$ \textit{optActionNN}.\textbf{action}( \textit{parentNode} )
		\State \textit{newState} $\gets$ \textbf{performAction}( parentNode, greedyAction )
		\State \textit{transCost} $\gets$ \textbf{getTransCost}( parentNode, newState )
		\State \textit{newNode} $\gets$ \textbf{createNodeObj}( \textit{newState}, \textit{randAction}, \hfill \\ 
											\hfill \textit{transCost} )
		\State \Return \textit{newNode}
	\end{algorithmic}
\end{algorithm}

\section{Experimentation} \label{experiments}
\noindent
In order to demonstrate efficacy of the proposed approach, three experiments were performed to test qRRT, namely, Differential-Drive Robot Navigation, Reactionless Manipulation of Space Robots and Two-Link Under-actuated Pendulum. Among the motion planning methods described in Sec. \ref{intro}, Anytime RRT was found to be most compliant with the requirements of non-holonomic planning, automatic cost learning and guaranteed optimality. Therefore, in this section, the results of qRRT have been compared with Anytime RRT. \par

\subsection{Motion Planning of Differential-Drive Robot over Regions with Varying Cost}
\noindent
We used this experiment to validate the qRRT algorithm and to analyze the influence of various parameters. Also, this offers a good graphical representation of the convergence of the qRRT method. This experiment consists of motion planning for a simulated differential drive robot over a terrain with varying cost regions. The cost of a point $[x,y]$ in the terrain is given by 
\begin{equation} \label{diffcost}
C(x,y) = -2 * abs( 20 * sin( 2\pi ( x/100) ) - y + 50 ) - 1 
\end{equation}
The lowest cost path is therefore a sinusoid with amplitude $20$, time period $100$ and centered around $y=50$. The robot has an axle width of 0.2m, wheel radius of 0.1m, and is restricted to paths tangential to the current arc being followed. \par
\begin{figure}[ht] 
	\centering
	\includegraphics[width=0.5\textwidth]{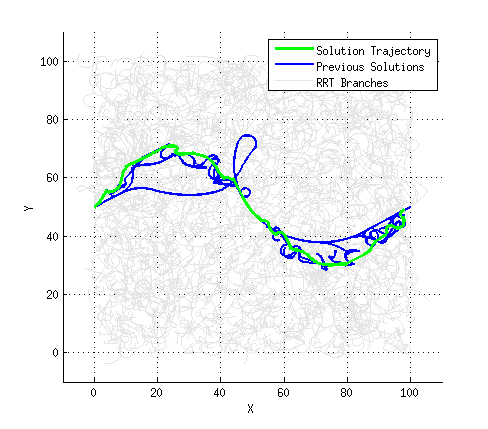}
	\caption{\small{Paths Generated by Differential Drive Robot using qRRT}}
	\label{fig:diffdrive}
\end{figure}

The experiment was conducted with goal bias of 1\%, quality bias increase at $0.3\%$ for every $10$ episodes and $\eta$ of $0.1$. The results of qRRT simulation after 300 episodes is given in Fig. \ref{fig:diffdrive}. It can be observed that even with minimal number of episodes, the solution path provided by qRRT has significantly converged towards the optimal path. The figure also includes a few of the earlier solution paths which show increased sampling around the low cost regions. The advantage of the incremental single-query approach is also evident in the workspace coverage by the branches of RRT. \par

\vspace{-5pt}
\vspace{-5pt}

As expected in RL implementations, solution cost improvement was observed when quality bias was gradually increased as the number of samples increased. Experiments with constant and high value of quality bias showed no significant improvement over successive episodes, implying lack of exploration. 

\subsection{Reactionless Maneuvering of a Dual-Arm Space Robot} \label{exp:reactless}
\noindent
Since space robots have a free-floating base, their dynamics are different from that of conventional earth-based systems. The coupling of the arms and the base of a space robot creates reaction forces and moments on the base whenever the arms execute a maneuver, causing the base to rotate and translate in accordance with the laws of conservation of linear and angular momenta. While small translations are acceptable, any change in attitude is undesirable from the point of view of communication and navigation. Consequently, several studies \cite{react1}, \cite{react2} have been conducted on ensuring that the reaction moment on the base is zero while executing maneuvers to ensure that the attitude of the base remains the same. This is termed reactionless maneuvering. \par
Constraints for reactionless manipulation are commonly written in terms of joint rate as \cite{react1}

\begin{equation}
\tilde{\boldsymbol{I}}_{bm} \dot{\boldsymbol\theta}=0.
\label{Eq3}
\end{equation}
where $\tilde{\boldsymbol{I}}_{bm}$ is the coupling inertia matrix between the base and arm, $\dot{\boldsymbol\theta}$ is the joint rates, and $\tilde{\boldsymbol{I}}_{bm} \dot{\boldsymbol\theta}$ is the coupling angular momentum. It is worth noting that the constraint for reactionless manipulation in Eq. \eqref{Eq3} is non-holonomic and this makes path planning a challenging problem. Any solution to Eq. \eqref{Eq3} is of the form
\begin{equation}
\mathit{\dot{\boldsymbol{\theta}}=\alpha_1\mathbf{v}_1+\alpha_2\mathbf v_2+...+\alpha_u\mathbf v_u},
\label{Eq6}
\end{equation}
where $\mathit{\alpha_i}$ are scalars, $\mathit{\mathbf v_i} \in R^{n\times 1}$ are vectors that span the null space, and $u$ is the dimension of the null space.

The $\mathit{\alpha_i}$ values are chosen to minimize the function
\begin{equation} ||\mathit{\dot{\boldsymbol{\theta}}-\dot{\boldsymbol\theta}_{req}}||^2 
\label{objfnalone}
\end{equation}
where
\begin{equation}
\begin{split}
\dot{\boldsymbol\theta}_{req}=-\lambda(\boldsymbol\theta_{current}-\boldsymbol\theta_{desired}),
\end{split}
\label{controleqn}
\end{equation}

$\lambda$ is a constant which controls the rate of convergence, and $\boldsymbol\theta_{current}$ is the set of current joint angles. Equation \eqref{controleqn} acts as a control law that tries to drive the system to a configuration $\boldsymbol\theta_{desired}$. Eqs. \eqref{Eq3} to \eqref{controleqn} allow us to evolve the state of the system in a reactionless manner using any sampling base planar. \par

Experimentation was performed on a 6-DOF planar dual arm space robot \cite{react2}. As Eq. \eqref{controleqn} is used for evaluation, the states contain 12 dimensions. 

\vspace{-5pt}
\begin{figure}[ht]
	\centering
	\includegraphics[width=0.4\textwidth]{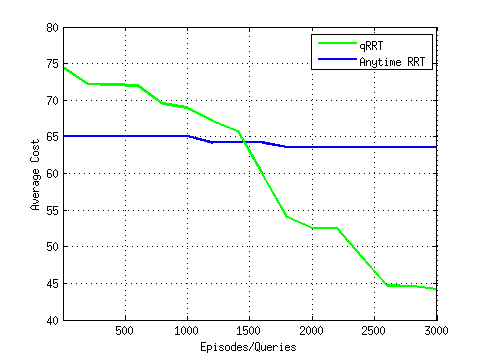}
	\caption{\small{Comparison of Anytime RRT and qRRT for Reactionless Manipulation}}
	\label{fig:reactless}
\end{figure}
\vspace{-5pt}

Since a steering function was not available for this system, minimization functions were used to determine the random extend that provided the best transition towards the random target. To improve execution time, the extend probabilities described in Sec. \ref{main_func} were adjusted to reduce the percentage of random extend operations. Exploration was performed mostly using randomly generated actions. Even in this setup, the randomly generated actions were able to sufficiently explore the configuration space and generate low cost trajectories. \par

Comparison was performed between qRRT and Anytime RRT. Anytime RRT was able to find a solution, but was unable to significantly improve the cost even after 3000 iterations. We believe this is due to use of Euclidean distance as the heuristic which cannot estimate the cost-to-go for the solution. When using qRRT as the solver, we used a fixed goal bias of 10\% and an increasing quality bias of 2\% for every 200 episodes. The cost of initial trajectories was similar to Anytime RRT with gradual improvement for the first ~1200 episodes. Significant reduction in cost was seen beyond this stage, coupled with increasing efficiency of the greedy trajectory returned by \textbf{getGreedyTraj}.  The results are shown in Fig.\ref{fig:reactless}. This proves superiority of the proposed approach in solving heavily constrained non-holonomic systems.

\subsection{Balancing Under-Actuated 2 Link Pendulum}
\noindent
The 2-link pendulum system consists of two arms connected by a powered joint and one of the links attached to a fixed point through an unactuated joint. The aim of the control is to achieve balance in the inverted position using application of torque in the actuated joint \cite{acrobot}. Experiments were conducted with both Anytime RRT (with Euclidean distance as heuristic) and qRRT. \par
An approximate steering function was used to perform random extend operations. Similar to Reactionless Manipulation, most of the exploration is performed using random actions. The qRRT algorithm was able to learn state values and greedy actions automatically and as sampling increased, improvements in both time taken to achieve balance and time taken to find a solution were observed. \par

\vspace{-5pt}
\begin{figure}[ht]
	\centering
	\includegraphics[width=0.4\textwidth]{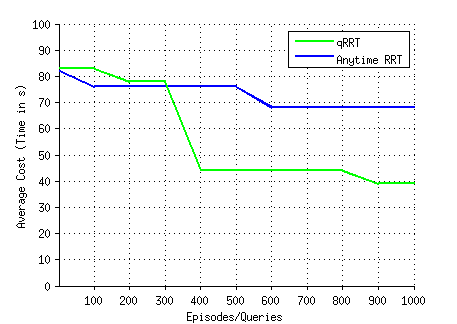}
	\caption{\small{Comparison of Anytime RRT and qRRT for Balancing of Under-actuated 2-Link Pendulum}}
	\label{fig:acrobot}
\end{figure}
\vspace{-5pt}

For qRRT, the quality bias was increased at 1\% for every 200 episodes and goal bias was fixed at 5\%. Time taken to achieve balance was chosen as the cost for this experiment. The cost improvement seen in qRRT was significantly higher than Anytime RRT. The comparative results are shown in Fig.\ref{fig:acrobot}.

\section{Conclusion} \label{conclusion}
\noindent
We present qRRT as an effective planner that can provide solutions for non-holonomic systems by combining exploratory search using RRT, cost-to-go estimation using RL and identification of low-cost paths using quality-biased exploitation of cost information. We have demonstrated that this method can be used in complex systems to identify and improve solution paths even in the absence of any cost information or known heuristic. The experiments also helped to establish the importance of the quality bias \% and the impact of its variation on the path solving process. \par
Future work in this method includes automatic tuning of qRRT parameters based on success/failure, cost improvement, and sample density. We would also like to identify methods to reuse the cost-to-go samples in the configuration space even when the system has undergone changes.

\clearpage

\end{document}